\documentclass[a4paper]{article}

\usepackage{INTERSPEECH2021}
\usepackage{graphicx}

\title{A Light-weight contextual spelling correction model for customizing transducer-based speech recognition systems\vspace{-0.5em}}
\name{Xiaoqiang Wang$^1$, Yanqing Liu$^1$, Sheng Zhao$^1$, Jinyu Li$^2$}
\address{
  $^1$Microsoft, China\\
  $^2$Microsoft, USA}
\email{\{xiaoqwa, yanqliu, szhao, jinyli\}@microsoft.com\vspace{-0.5em}}

\begin{document}

\maketitle
\begin{abstract}
  It’s challenging to customize transducer-based automatic speech recognition (ASR) system with context information which is dynamic and unavailable during model training. 
  In this work, we introduce a light-weight contextual spelling correction  model to correct context-related recognition errors in transducer-based ASR systems. 
 We incorporate the context information into the spelling correction model with a shared context encoder and use a filtering algorithm to handle large-size context lists.  
  Experiments show that the model improves baseline ASR model performance with about 50\% relative word error rate reduction, which also significantly outperforms the baseline method such as contextual LM biasing. The model also shows excellent performance for out-of-vocabulary terms not seen during training. 
\end{abstract}
\noindent\textbf{Index Terms}: speech recognition, contextual spelling correction, contextual biasing

\section{Introduction}

Recently, end-to-end (E2E) modeling becomes the most significant trend for automatic speech recognition (ASR). Because of its streaming nature, transducer models \cite{Graves-RNNSeqTransduction}  including recurrent neural network (RNN) Transducer (RNN-T) \cite{OTF_rescore3_0, Li2019RNNT, saon2020alignment, zeyer2020new} and Transformer Transducer \cite{yeh2019transformer, zhang2020transformer, chen2020developing}. While achieving similar or even better recognition accuracy than hybrid models \cite{OTF_rescore3_0, Li2020Developing}, transducer models are facing lots of practical challenges. Customizing transducer models with dynamic context is one of the most challenging tasks. 
For example, in an intelligent personal assistant, ``\textit{call} $<$\textit{PersonName}$>$'' is customized with the user’s contact list. Such contexts are dynamic, personalized, or occasion related. More examples include names, songs, playlist, location, etc. These contexts have not been observed during training, therefore transducer models usually perform poorly when recognizing these contexts. 

There are several studies trying to improve the accuracy of customized transducer models. The first approach is to integrate the contextual information in the decoding process by rescoring the output probabilities with a contextual language model (LM) constructed from context phrases, referred to as shallow fusion or on-the-fly rescoring \cite{OTF_rescore1, OTF_rescore2, OTF_rescore3, OTF_rescore4}. The rescoring method itself is a one-pass process, which also relies heavily on the ASR model output distributions and the weights should be tuned carefully in different using scenarios. The second approach, contextual RNN-T \cite{jain2020contextual}, generally encodes contextual information into the transducer model and regards the contexts as additional model input apart from audio features. However, this method may not be scaled well with a large biasing list. Furthermore, because contextual RNN-T changes the baseline model structure, it is more expensive in training and inference, and may result in performance degradation as noted in the work from the same organization \cite{le2021deep}. Although the deep shallow fusion \cite{le2021deep} can achieve reasonable performance for customization, it significantly degrades the performance on general sets.



In this work, we propose a novel contextual biasing method which leverages contextual information by adding a contextual spelling correction (CSC) model on top of the transducer model. To consider contextual information during correction, a context encoder which encodes context phrases into hidden embeddings is added to the spelling correction model \cite{SC_1, SC_2}, the decoder of the correction model then attends to the context encoder and text encoder by attention mechanism \cite{attention_mechanism}. The proposed model is a standalone correction model which does not change the original transducer model structure, hence there is no performance degradation risk for the baseline ASR model. It’s convenient for the proposed method to be applied in different domains by just changing the CSC model without retraining the original ASR model. To address the problem of general domain regression and large context list, we propose a filter mechanism to trigger the correction process and control the context list size based on ASR hypotheses. Experiments show that the proposed CSC model significantly improves the ASR accuracy of RNN-T models in domains with contextual phrases such as personal names and outperforms the baseline biasing methods. The model also shows excellent performance for out-of-vocabulary (OOV) terms not seen during training. With the help of teacher-student learning \cite{li2014learning, knowledge_distill} and quantization \cite{quant}, the model can be compressed to a small size, which makes it easy to be deployed. 

\section{Related Work}

\subsection{Contextual LM}

Contextual LM is adopted as one of our baseline models by following the shallow-fusion end-to-end contextual biasing method proposed by Zhao et.al \cite{OTF_rescore4}. Given the audio input $x$, shallow fusion interpolates the transducer ASR model with an external contextual LM:\vspace{-0.1em}
\begin{equation}
{y}^* = \mathop{\arg\max}_{y}{logP(y|x)}+\lambda{logP_{c}(y)}   
\end{equation}\vspace{-0.1em}
Where $P$ and $P_c$ are the probabilities from baseline ASR model and contextual LM, $\lambda$ is a tunable parameter. The contextual LM, $min(det({S}\circ{G}))$, is obtained by compiling the list of biasing phrases into an n-gram weighted finite state transducer (WFST) \cite{WFST} $G$, Where $G$ is left composed with a “speller” FST $S$ to transduce wordpieces \cite{wordpiece} into the corresponding word. To address the concern that biasing hurts the accuracy of utterances without contexts, activation prefix \cite{OTF_rescore4} is also integrated and biasing is activated only when it’s preceded by prefix. 

\subsection{Bias encoder}

There have been growing interests in incorporating contextual information into E2E ASR model with bias encoders.
Pundak et al. proposed CLAS model \cite{CLAS} which adds a context bias encoder to LAS \cite{LAS} by taking context phrases as input. The decoder calculates attention with both the audio encoder and bias encoder, and the obtained attention vectors from the two encoders are then concatenated to generate the final attention. Similar idea was applied to RNN-T as contextual RNN-T \cite{jain2020contextual}. However, this method may not be scaled well when the context list is large, and it also complicates the training and inference.


\subsection{Spelling correction}

Spelling correction \cite{SC_1, SC_2} in ASR system aims to correct recognition errors in ASR hypotheses, which is shown to be effective in model performance improvement. There is also some work targeting on proper noun recognition problem \cite{G2G}. Generally, spelling correction is more like machine translation \cite{machine_translation} which “translates” ASR hypotheses with error terms into the correct one. Different from traditional spelling correction, we additionally correct ASR hypotheses with contextual information, which takes dynamic biasing phrases for customization. 

\section{Contextual spelling correction}

\subsection{Model structure}

\begin{figure}[h]
  \centering
  \includegraphics[width=\linewidth]{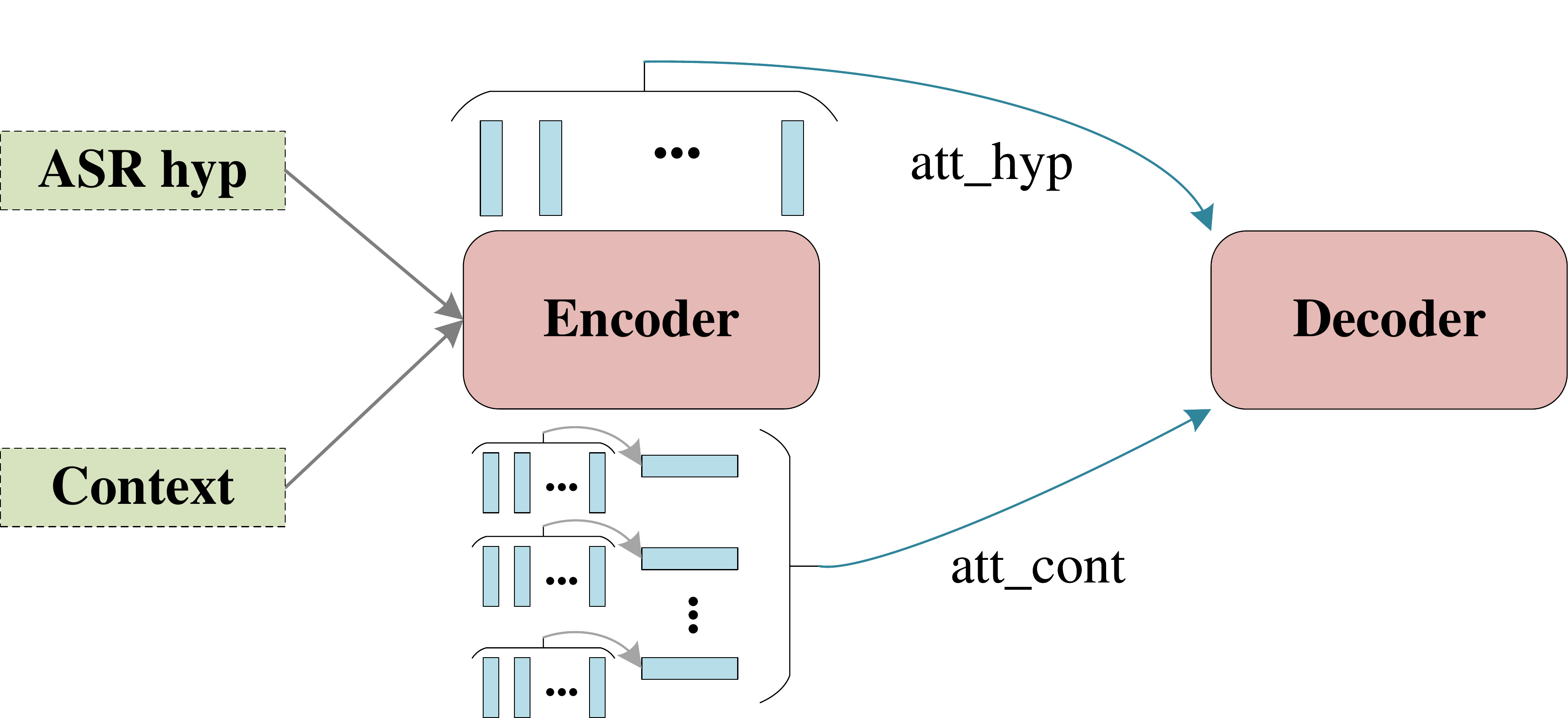}
  \caption{Contextual spelling correction (CSC) model structure.}
  \label{fig:CSC}
\end{figure}

The proposed model is a seq2seq \cite{seq2seq} model with a text encoder, a context encoder, and a decoder, which takes ASR hypothesis as the input to text encoder, context phrase list as the input to context encoder. The context encoder encodes context phrases as context embeddings, the decoder attends to the output of both encoders, obtaining information from ASR hypotheses hidden states and context embeddings to correct contextual misspelling. All the components are transformer-based \cite{attention}, composed by pre-LayerNorm \cite{layernorm, GPT2}, self-attention, encoder-decoder attention and feed-forward layer (FFN). The parameters of the two encoders are shared, as shown in figure \ref{fig:CSC}. In the following, we describe the key features of the proposed model.

\subsubsection{Context encoder}
Different from general spelling correction model, the contextual information is incorporated with this context encoder as embeddings, which is generated by averaging the hidden representations of the context phrase as a single vector. At each decoding step, the query vector also pays attention to context phrase embeddings, the attention of text encoder and context encoder are then added up to generate the final attention during decoding.

\subsubsection{Light Weight model}

To enable on-device application, we reduce the model size and improve inference efficiency with the following methods:

\begin{itemize}
\item Parameter sharing.
Parameters are shared between the text encoder and context encoder by using a single encoder network for feature extraction of both ASR hypotheses and contexts. This design comes from: 1) ASR hypotheses and context phrases are both transcriptions that could be processed by the same network; 2) In many cases such as domain with user names, the context phrase list for training is not that large to cover all the possible word tokens or patterns, using the same network makes the context encoder benefit from text encoder; 3) Use a single encoder network makes the model smaller. Furthermore, the embeddings of encoder and decoder are also shared.
\item Teacher-student learning \cite{li2014learning, knowledge_distill} and quantization \cite{quant}. Both methods are used to reduce the model size.
\end{itemize}



\subsection{Training}

\subsubsection{Text to speech data}

In most scenarios, it is time and cost consuming to get enough speech data with context phrases and label them. Generating text to speech (TTS) audio using preprocessed transcriptions with context phrases is an efficient and low-cost way to augment the training data. 

We first collect a number of sentence patterns with contexts (in our case, person names), such as "\textit{call} $<$\textit{PersonName}$>$", "\textit{do I have any mails from} $<$\textit{PersonName}$>$ \textit{today}", etc, and combine these sentence patterns with a prepared context list randomly to get training scripts. Then we use TTS to generate audio and run through the RNN-T model to get ASR hypotheses. The input reference, context phrases and generated ASR hypotheses are finally paired for CSC model training.

\subsubsection{Data augmentation}

Due to the sparsity of transcripts with contexts, the collected sentence patterns is limited, which will bring about overfitting on these sentence patterns. There are two approaches to deal with this problem. The first method is to use some NER \cite{NER_1, NER2} tools to filter out sentences with contexts from large amount of data. The second one is to randomly replace words in general sentence with ASR hypotheses-reference pairs of prepared context phrases to generate training pairs artificially, e.g, \{$x$: \textit{you are Bob}, $y$: \textit{you are Bobby}\} may be generated from "\textit{you are right}" by replacing word "\textit{right}" with pairs \{\textit{Bob}, \textit{Bobby}\}. This method is simpler but may destroy the grammar to some extent. To make the data preparation process simpler, we tested the second method. From the training results we found it does not influence the model performance much. 

\subsubsection{Context setup}

During training, each utterance has one or two ground-truth context phrases. We pool all the context phrases from all utterances in the batch together and use this to sample the context phrase list for each utterance, a mask is used to distinguish the sampled context phrase list among these utterances. The size of context phrase list $N_c$ for each utterance is randomly sampled from a uniform distribution $[1, N_{cmax}]$, where $N_{cmax}$ is max context list size. To take care of the general sentences without context or cases that the ground-truth context doesn't appear in context list, we keep the proportion of training samples without ground-truth context to be $1-P_{cont}$. What’s more, similar to \cite{CLAS, phoebe, keyword}, a token $<$/cont$>$ is added after the context phrase $c$ in reference $y$. A model input example is described as:

ASR hyp $x$: \textit{Who is Alyssa Friedman.}

Contexts $c$: \textit{Samira, Trump, Aliza Friedman, ..., Joe Biden}

Reference $y$: \textit{Who is Aliza Friedman $<$/cont$>$.}

\subsubsection{Teacher-student learning}

After the basic CSC model is trained, teacher-student learning \cite{li2014learning} is also adopted to further reduce the model size and improve the inference efficiency, which enables the model to be used on device. The loss function of the student model is:
\begin{equation}
L = {\alpha}L_{soft}+(1-\alpha)L_{hard}   
\end{equation}
\begin{equation}
L_{hard} = H(y_{S}, y)   
\end{equation}
\begin{equation}
L_{soft} = D_{\rm KL}\left({\rm softmax}(\frac{y_{S}}{T}), {\rm softmax}(\frac{y_{T}}{T})\right) \cdot T^2
\end{equation}
Where $L_{hard}$ is cross-entropy loss of student model output $y_{S}$ and reference $y$, $L_{soft}$ is KL-divergence of $y_{S}$ and teacher model output $y_{T}$. $T$ is a temperature parameter, and $\alpha$ is a weight to determine the proportion of $L_{hard}$ and $L_{soft}$.

\subsection{Inference}

During inference, the ASR hypotheses is first filtered by a pattern classifier, and the context list is filtered by a relevance ranker and a preference ranker to address the large size context list issue in existing solutions like CLAS. Then the processed data is fed into the CSC model to correct contextual errors.

\subsubsection{Pattern classifier}

The pattern classifier aims to classify the input ASR hypotheses into ``to be corrected'' or ``not to be corrected'' in order to avoid the regression of general sentences that we do not want to bias. 
For simplicity, we adopt a rule-based classifier which is generated from training scripts and considers the sentence patterns (e.g., sentences with "\textit{call}", "\textit{forward ... to}", etc.) to determine when to trigger our CSC model. 

\subsubsection{Relevance ranker (rRanker)}

To avoid performance degradation and increase inference efficiency when the context list is extremely large, a relevance ranker is used to constrain context list size based on the ASR hypotheses. For sentence $s$, we calculate the relevance ranker weight of a certain context phrase $c_j$ as
\begin{equation}
W_r^j = -\frac{\min_{i}({\rm editdistance}(c_j, s_i))}{{\rm len}(c_j)}
\end{equation}
where $s_i$ is a segment cut off from input text $s$ with the same length of the context phrase $c_j$ begin from the $i$-th word.

\subsubsection{Preference ranker (pRanker)}

In most scenarios, there is additional preference knowledge for us to determine the prior distribution of the context list. For example, in the personal assistant scenario, the context preference of each contact in the user’s contact list can be determined as the query frequency of this contact. To utilize this information, we also introduce a rank weight $W_p$ based on user’s preference to preselect context together with relevance ranker weight $W_r$. The final selected context list can be described as:
\begin{equation}
c_1, c_2, ..., c_k = {\arg}{\rm topk}(\alpha_{p}W_{p}+(1-\alpha_{p})W_{r})
\end{equation}
Where ${\alpha}_p$ determines the proportion of relevance ranker weight $W_r$ and preference rank weight $W_p$. 

\subsubsection{Decode with CSC}

For each utterance, we get top $N_{asr}$ ASR hypotheses $\{H_1, H_2, ..., H_{N{_{asr}}}\}$ which may contain contextual spelling errors. Each hypothesis $H_i$ is then fed into the CSC model with the pre-selected context list and the corresponding CSC hypotheses $\{H_{i1}, H_{i2}, ..., H_{iN_{csc}}\}$ is generated by beam search mechanism. The final decoding results are obtained by ranking these $N_{asr}\times{N_{csc}}$ hypotheses:
\begin{equation}
H^* = \mathop{\arg\max}_{H}{\lambda_{asr}{logP_i}+\lambda_{csc}{logP_{ij}}}
\end{equation}
where $\lambda_{asr}$ and $\lambda_{csc}$ are the weights for ASR and CSC scores.

\section{Experiment}


Our experiment targets on correcting person name spelling errors for an English-speaking personal assistant scenario in which name contexts are occasionally triggered. To generate training scripts with contexts, we first collected 512 sentence patterns with name token that frequently appear in this scenario. Then 1 million (M) training scripts are constructed by randomly combining these sentence patterns with 1M names which are generated from 48 thousand (K) name words. These scripts are used to generate TTS audio as described in section 3.2.1. We use Fastspeech AM \cite{fastspeech} + lpcnet vocoder \cite{LPCnet} as the TTS model. To augment the training set, we also used another 26M in-house general scripts which are also from this scenario following the approach described in section 3.2.2.

We use two test sets to evaluate the model performance, including a test set with person names (denoted as name set) and a general set without names. The name set contains 11K utterances. The name list size (context phrase list size) $K_r$ of each utterance is 1509 on average. This is much larger than the size in \cite{jain2020contextual, le2021deep}, indicating our setup is more challenging. The general set contains 8K utterances without names.

\subsection{Experimental Setup}

During training, we select $batch\_size=250$, $N_{cmax}=100$ and $P_{cont}=0.8$, which means the max size of context phrase list of an utterance is 100 in training process, and there are about 80$\%$ utterances with context inside each batch. Based on the above settings, we trained a teacher model, and then distilled it to a student model using the same training scripts, the distillation parameters are set to be $T=1$, $\alpha=0.9$. The detailed CSC model parameters are listed in Table \ref{tab:model_parameters}. During inference, the size of filtered context list $K_f$ after rRanker and pRanker is set to be 80 to get consistent with training. We also take $N_{asr}=4$, $N_{csc}=1$ and $\alpha_{p}=0.3$ during CSC decoding.

\begin{table}[t]
  \caption{CSC Model parameters}
  \label{tab:model_parameters}
  \centering
  \resizebox{\linewidth}{!}{%
  \begin{tabular}{cccccc}
    \toprule
    \textbf{Model} & layers & d\_model & heads & d\_FFN & Params(M)  \\
    \midrule
    \textbf{Teacher} & $6$ & $512$ & $8$ & $2048$ & $55.4$  \\
    \textbf{Student} & $3$ & $192$ & $4$ & $768$ & $5.2$ \\
    \bottomrule
  \end{tabular}}
\end{table}


\subsection{Performance}

Table \ref{tab:rnnt_version} shows the CSC model performance on three RNN-T models. 
All these RNN-T models have 1600 LSTM \cite{LSTM} memory cells and the output is projected to 800. The encoder has 6 layers while the prediction network has 2 layers LSTM. The joint network outputs a vector with dimension 640. The feature is 80-dimension log Mel filter bank for every 10 milliseconds (ms) speech. Three of them are stacked together to form a frame of 240-dimension input acoustic feature to the encoder network. The output targets are 4 thousand word-piece units. As described in \cite{Li2020Developing}, all these RNN-T models were trained with 65 thousand hours of anonymized transcribed data with personally identifiable information removed, differing only at how many future frames were observed at each encoder layer. 

The CSC model was trained using the v1 RNN-T model. We report both WER and relative WER reduction (WERR) from the baseline RNN-T model for the proposed RNN-T+CSC method in Table \ref{tab:rnnt_version}. We can see that final model performance does not change too much among different RNN-T models, which indicates the robustness of the proposed method. 

\begin{table}[t]
  \caption{Model perf with ASR version change on name set}
  \label{tab:rnnt_version}
  \centering
  \resizebox{\linewidth}{!}{%
  \begin{tabular}{llll}
    \toprule
    \textbf{Model}  & \textbf{v1}  & \textbf{v2} & \textbf{v3}   \\
    \midrule
    RNN-T      & $28.9$ & $26.8$ & $30.4$   \\
    RNN-T+CSC  & $14.5(49.8\%)$ & $13.3 (50.4\%)$ & $16.1 (47.0\%)$     \\
    \bottomrule
  \end{tabular}}
\end{table}

\begin{table}[t]
  \caption{Model performance, WER(WERR)}
  \label{tab:model_perf}
  \centering
  \resizebox{\linewidth}{!}{%
  \begin{tabular}{lll}
    \toprule
    \textbf{}  & \textbf{Name set}  & \textbf{General set}   \\
    \midrule
    RNN-T      & $30.4$         &  $13.4$  \\
    RNN-T+contextual LM  & $22.2 (27.0\%)$  & $13.6 (-1.5\%)$    \\
    RNN-T+CSC (w/o pRanker) & $17.4 (42.8\%)$  & $13.8 (-3.0\%)$      \\
    RNN-T+CSC  & $16.1 (47.0\%)$  & $13.7 (-2.2\%)$     \\
    \bottomrule
  \end{tabular}}
\end{table}

In Table \ref{tab:model_perf}, we compare the performance of CSC with the baseline RNN-T model and the RNN-T model with contextual LM (on-the-fly rescoring) using the v3 model. The proposed model achieves 47.0$\%$ WERR while the model based on contextual LM achieves 27.0$\%$ WERR, with only slight regression on the general set. 
The context preference ranker which acts as prior knowledge improves the model performance. 

Table \ref{tab:teacher_student} shows performance of CSC teacher model and 8-bit quantized student model on CPU (Xeon E5-2690 v4 @ 2.60GHz) with single thread. Both models are in ONNX format. The student model without rRanker and pRanker feeds the full context list into CSC model without context preselection operation. Which shows with teacher-student learning, quantization and context filtering algorithm, the overall runtime performance of the model is greatly improved.

\begin{table}[t]
  \caption{Runtime performance for teacher/student}
  \label{tab:teacher_student}
  \centering
  \resizebox{\linewidth}{!}{%
  \begin{tabular}{cccc}
    \toprule
    \textbf{Model}  & \textbf{Size(MB)}  & \textbf{WER} & \textbf{Latency}   \\
      &   &  & \textbf{(ms/utterance)}   \\
    \midrule
    Teacher      & $223.0$ & $15.6$ & $793.6$   \\
    Student  & $9.5$ & $16.1$ & $132.3$     \\
    Student w/o r(p)Ranker  & $9.5$ & $19.3$ & $1846.7$     \\
    \bottomrule
  \end{tabular}}
\end{table}

\subsection{OOV contexts}

The result of out of vocabulary (OOV) items is also important for contextual biasing, which indicates the model performance in real online scenario. We define OOV rate as the fraction of contexts in test set that are not seen during training and measure from three aspects: full name OOV rate (e.g., the full name “\textit{Joe Biden}” not in training set), one name word OOV rate (e.g., “\textit{Joe}” in training set while “\textit{Biden}” is not in training set), and all name words OOV rate (e.g., neither “\textit{Joe}” nor “\textit{Biden}” is in training set), denoted as OOV1, OOV2, OOV3. These OOV rates for the name set are 54$\%$, 15.9$\%$ and 5$\%$, respectively. Table \ref{tab:oov} shows model performance for OOV items and its complementary set, denoted as OOV\_C. The CSC model improves more on OOV items than on context phrases covered in training. One of the reasons is that the baseline WER for the OOV items of the baseline RNN-T model is larger due to more contextual spelling errors. 

\begin{table}[t]
  \caption{Model performance for OOV items, WER(WERR)}
  \label{tab:oov}
  \centering
  \begin{tabular}{lcc}
    \toprule
    \textbf{OOV} & \textbf{RNN-T}  & \textbf{RNN-T+CSC}  \\
    \midrule
    OOV1     & $37.2$  & $16.5 (55.6\%)$  \\
    OOV1\_C     & $22.5$  & $15.4 (31.6\%)$ \\
    OOV2     & $43.3$  & $19.1 (55.9\%)$ \\
    OOV2\_C     & $27.3$  & $15.3 (44.0\%)$ \\
    OOV3     & $37.1$  & $16.9 (54.4\%)$  \\
    OOV3\_C     & $30.2$  & $16.0 (47.0\%)$ \\
    \bottomrule
  \end{tabular}
\end{table}

\subsection{Context list size}
\begin{figure}[h]
  \centering
  \includegraphics[width=\linewidth]{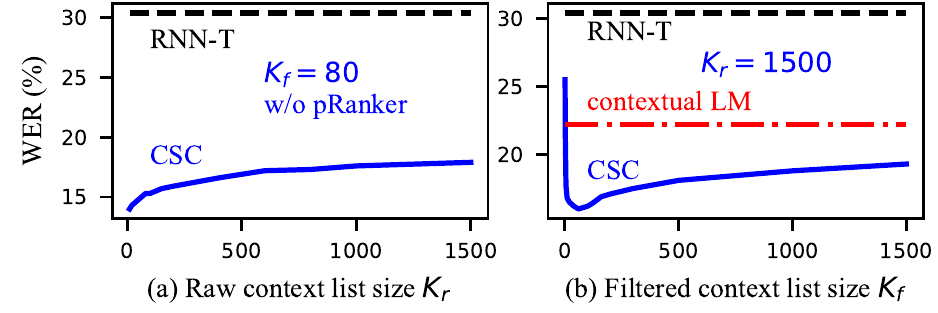}
  \caption{Influence of context list size to model performance}
  \label{fig:contsize}
\end{figure}

During inference, the input context list size affects the model performance. We have two context list sizes: the raw context list size $K_r$ before filter and the filtered context list size $K_f$. Figure \ref{fig:contsize}(a) illustrates the influence of $K_r$ to model performance, which shows the WER increases when the raw context list size becomes larger. Figure \ref{fig:contsize}(b) illustrates the influence of $K_f$ to model performance, we can see the WER first drops and then increases when $K_f$ becomes larger, this is because when $K_f$ is set to small, the filter may miss the ground-truth context. There is always a best value for filtered context list size $K_f$, which depends on the context filter and context list distribution.

\section{Conclusions}

In this work, we introduce a novel light-weight CSC model for customizing the transducer-based ASR systems. The context information is integrated with a context encoder to a spelling correction model. Novel filtering algorithms are designed to deal with the large size context list.  Encoder sharing, teacher-student learning, and quantization are used to achieve low runtime cost. Experiments show that the CSC model significantly outperforms the contextual LM biasing method, with around 50\% relative WER reduction from the general RNN-T model on the name recognition set, while with only slight regression on the general set. The model also shows excellent performance for OOV terms not seen during training. Although the proposed method is verified with RNN-T in this study, it can also be applied to any transducer models.

\bibliographystyle{IEEEtran}

\bibliography{main}


\end{document}